\documentclass[a4paper,fleqn]{cas-dc}

\usepackage[numbers]{natbib}
\usepackage[commandnameprefix=always,final]{changes} 

\definechangesauthor[name={Qianxue Zhang}, color=blue]{R1}

\begin{document}
\let\WriteBookmarks\relax
\let\printorcid\relax
\def\floatpagepagefraction{1}
\def\textpagefraction{.001}
\shorttitle{}
\shortauthors{}

\title [mode = title]{Toward Vibe Medicine: A Self-Evolving Multi-Agent Framework for Clinical Decision Support}

\author[1,2,3,4]{Qianxue Zhang}
\fnmark[1] 
\affiliation[1]{organization={Medical AI Lab},
                addressline={The First Hospital of Hebei Medical University}, 
                city={Shijiazhuang},
                postcode={050000}, 
                state={Hebei},
                country={China}}

\author[1,2,3,4]{Yiming Ren}
\fnmark[1] 

\affiliation[2]{organization={Hebei Provincial Medical Artificial Intelligence Research Institute},
                city={Shijiazhuang},
                postcode={050000}, 
                state={Hebei},
                country={China}}

\affiliation[3]{organization={Hebei Provincial Engineering Research Center for AI-Based Cancer Treatment Decision-Making},
                addressline={The First Hospital of Hebei Medical University}, 
                city={Shijiazhuang},
                postcode={050000}, 
                state={Hebei},
                country={China}}

\author[1,2,3,4]{Shihuan Qin}
\affiliation[4]{organization={State Key Laboratory of Neurology and Oncology Drug Development},
                state={Nanjing},
                country={China}}

\author[1,2,3]{Xiao Zhang}
\author[1,2,3]{Liao Zhang}

\author[1,2]{Jinyang Huang}

\author[5]{Zhengliang Liu}
\affiliation[5]{organization={School of Computing, University of Georgia},
                addressline={415 Boyd Research and Education Center}, 
                city={Athens},
                postcode={30602}, 
                state={GA},
                country={USA}}

\author[6]{Chenbin Liu}
\affiliation[6]{organization={Department of Radiation Oncology, National Cancer Center/National Clinical Research Center for Cancer/Cancer Hospital and Shenzhen Hospital, Chinese Academy of Medical Sciences and Peking Union Medical College},
                city={Shenzhen},
                country={China}}

\author[7]{Hongying Feng}
\affiliation[7]{organization={College of Mathematics and Physics, China Three Gorges University},
                city={Yichang},
                state={Hubei},
                country={China}}

\author[8]{Jingyuan Chen}
\affiliation[8]{organization={Department of Radiation Oncology, Mayo Clinic},
                addressline={5881 E. Mayo Blvd.}, 
                city={Phoenix},
                postcode={85054}, 
                state={AZ},
                country={USA}}

\author[8]{Yuzhen Ding}

\author[5]{Weihang You}
\author[5]{Hanqi Jiang}
\author[5]{Yi Pan}
\author[5]{Yifan Zhou}
\author[5]{Junhao Chen}
\author[5]{Lifeng Chen}

\author[8]{Wei Liu}

\author[5]{Tianming Liu}
\cormark[1]

\author[9,2,10]{Zengren Zhao}
\affiliation[9]{organization={Gastrointestinal Disease Diagnosis
and Treatment Center, The First Hospital of Hebei Medical University}, 
                city={Shijiazhuang},
                postcode={050000}, 
                state={Hebei},
                country={China}}

\affiliation[10]{organization={Department of General Surgery, The First Hospital of Hebei Medical University}, 
                city={Shijiazhuang},
                postcode={050000}, 
                state={Hebei},
                country={China}}

\cormark[1]

\author[1,2,3]{Lian Zhang}
\cormark[1]

\cortext[cor1]{Corresponding authors: lianzhang@hebmu.edu.cn (L. Zhang); zhaozengren@hebmu.edu.cn (Z. Zhao); tliu@uga.edu (T. Liu)}
\fntext[fn1]{Co-first authors: Qianxue Zhang and Yiming Ren.}


\begin{abstract}
In recent years, the advances of large language models and autonomous agents have revolutionized the healthcare field, facilitating diagnosis and improving treatment results. However, most existing AI systems rely on pre-trained knowledge and predefined pipelines, which struggle to learn dynamically from the interactive chat session history that contains patient outcomes and past failures. To address this limitation, we propose VIBEMed, a multi-agent framework with a built-in self-evolution mechanism and architecture-level safety sandbox for robust clinical decision support. The system integrates three specialized agents, including a Clinical Diagnostic Agent (CDA) for hypothesis generation, a Therapeutic Execution Agent (TEA) for treatment planning, and a Clinical Evolution Manager Agent (CEMA) that distills longitudinal clinical feedback into reusable knowledge, transforming multimodal patient information into personalized medical decisions. Through self-evolution mechanism, the framework enables iterative updates across memory, model behavior, and decision strategies, allowing the system to improve over time. Experimental results show that VIBEMed demonstrates superior performance through its evolving mechanism in complex clinical cases, particularly in tasks that require integrated decision-making and longitudinal planning. The framework also supports reliable end-to-end decisions in challenging scenarios such as oncology treatment planning, highlighting its feasibility in real-world clinical contexts. Overall, VIBEMed provides a practical path beyond static AI systems toward adaptive, experience-driven clinical decision support, demonstrating the value of combining multi-agent collaboration with continuous evolution for advancing precision medicine.

\end{abstract}


\maketitle

\section{Introduction}
The rapid evolution of artificial intelligence is redefining the computational domain, transitioning from traditional natural language processing (NLP) techniques to large-scale models with trillions of parameters, while further developing into agents with independent thought and task execution capabilities. By integrating intent understanding, complex reasoning, and tool invocation, general-purpose large language models (LLMs) and intelligent agents drastically enhance information processing efficiency and demonstrate exceptional productivity in complex workflows such as software development and scientific research \cite{hou2024large}. The advancement from simple text generation to multi-step task planning marks a fundamental shift, with artificial intelligence evolving from a passive tool into an active driver of productivity.

Moreover, the AI technologies and applications have reformed the traditional healthcare sector, benefiting diagnostic process and optimizing administrative workflows \cite{thirunavukarasu2023large}. Recent studies indicate that multi-agent systems and advanced models can effectively assist in analyzing imaging reports through automated segmentation and 3D reconstruction, support complex clinical decision-making through multimodal data integration, and enable personalized patient treatment through chain-of-thought (CoT) reasoning and attention mechanisms, thus significantly improving healthcare efficiency and reducing the burden on medical professionals \cite{li2025care, peng2024integration}. In the meantime, the standardization of healthcare management process can be further enhanced by leveraging intelligent scheduling and resource allocation tools.

Despite their broad application prospects, existing medical AI systems still face critical bottlenecks \cite{zheng2025large, kelly2019key}. Most current architectures rely heavily on static pre-trained weights and preset execution workflows, with their core capabilities primarily acquired through offline training, which limits their ability to be continuously updated during interactive sessions \cite{shi2025continual}. Although modern LLMs can leverage attention mechanisms \cite{vaswani2017attention} to retain contextual information within a conversation, such memory remains
short-term, and it cannot persist beyond individual sessions or lead to updates in the model's internal parameters. Consequently, models struggle to effectively acquire and integrate new knowledge from dialogue histories, longitudinal patient prognosis, and prior treatment failures, resulting in poor performance in unseen clinical scenarios. Without feedback-based continuous learning mechanisms, models are confined to fixed pattern matching and unable to evolve overtime.

To address the knowledge gap, several research has introduced retrieval-augmented generation (RAG) \cite{lewis2020retrieval} and medical knowledge graphs to incorporate external knowledge into LLMs \cite{singhal2023large}. While these approaches improve access to up-to-date medical information, they are still limited in capturing the complex temporal and causal relationships in patient data \cite{peng2025graph}. Besides, complex clinical diagnosis and treatment typically involve multiple steps and coordinated decision-making, which cannot be adequately supported by traditional retrieval mechanisms. Existing methods mainly rely on semantic similarity or static graph structures, which often produces sparse or irrelevant information that is difficult to integrate into coherent clinical decisions. These limitations highlight the need for more structured and adaptive frameworks to better model real-world clinical workflows.

In addition, intelligent healthcare applications face critical challenges related to security, privacy, and ethics \cite{shen2023chatgpt, haltaufderheide2024ethics}. Since medical data is highly sensitive, fully automated agents without robust security measures are vulnerable to cyberattacks and privacy breaches, posing risks to protected health information \cite{zhong2025considerations}. In order to protect patient privacy, LLMs are advised to be deployed within isolated hospital local networks to ensure data security, but given the limited availability of open-source medical LLMs and differences in patient demographic distributions as well as institution-specific clinical protocols, the models tend to exhibit degraded performance when applied in real-world clinical environments \cite{chen2024integration}. Without the ability to fine-tune or evolve, this lack of local adaptability makes it difficult to ensure safety and clinical efficacy in complex interactions.

To overcome the above limitations, a self-evolving architecture based on context awareness and historical memory have attracted increasing attention. Existing research has introduced the concepts of vibe coding \cite{meske2025vibe} and vibe research \cite{lyu2026evoscientist}, where agents can start from simple prompts and gradually improve by learning from interaction histories and experimental outcomes, similar to how human cultivate skills through experience. Eventually, agents successfully develop the ability to handle complex tasks through self-evolution \cite{fang2025comprehensive}. 
Inspired by these advances and the experience-driven nature of clinical practice, we introduce the concept of vibe medicine, a paradigm in which medical AI systems continuously evolve by learning from real-world interactions, accumulating clinical experience, and refining their decision-making over time. Therefore, VIBEMed (Versatile Intelligent Behavior-Evolving Medical framework) is proposed as an end-to-end self-evolving multi-agent system, which enables continuous learning without relying on high-quality prompt engineering or constant human supervision. By introducing multi-role collaboration and a mechanism for persistent clinical memory, it endows medical agents with the ability to self-correct and continuously evolve through interaction. The core innovations of this paper encompass the following aspects:
\begin{itemize}
    \item An innovative multi-agent collaborative framework for clinical decision-making: The proposed architecture consists of three specialized agents, including a Clinical Diagnostic Agent, a Therapeutic Execution Agent, and a Clinical Evolution Management Agent. This framework simulates real-world treatment workflow, enabling end-to-end automation from multimodal patient data analysis to personalized treatment planning.
    \item A novel three-level self-evolution mechanism: VIBEMed proposes autonomous evolution across memory, model and code, balancing rapid adaptation with long-term stability. At the memory level, the system distills successful experiences and failure cases from historical interactions into a persistent memory for real-time improvements. At the model level, it constructs reflection datasets and performs staged updates via supervised fine-tuning (SFT) and direct preference optimization (DPO), enhancing the LLM’s core capabilities under controlled validation. At the code level, the framework refines its structure through optimization of agent interaction logic and deployment of new functional modules based on user feedback. This unified mechanism enables continuous and safe adaptation without manual intervention.
    \item An architecture-level safety sandbox for clinical reliability: VIBEMed enforces safety through system-level constraints rather than relying on agent behavior or manual oversight. The framework adopts isolated execution environments to separate development, validation, and production stages, preventing unverified updates from reaching clinical use. In addition, a session-level memory isolation mechanism ensures strict separation of patient data through scoped access control across all memory tiers. System operations are further secured with structured validation, user acknowledgment for new functionalities, and immutable audit logging for traceability. This design enables autonomous operation while maintaining the safety and reliability required in clinical settings.

\end{itemize}

\section{Material and methods}
In this section, we present the detailed design of the VIBEMed framework. First, we provide a comprehensive system overview, highlighting the key building blocks and mechanisms involved. Next, the multi-agent collaborative network is demonstrated, followed by a detailed explanation of the three-level evolution mechanism. Finally, we introduce the architecture-level safety sandbox designed to ensure secure and reliable system operation.

\subsection{System overview}
\begin{figure*}[!htbp]
	\centering
	\includegraphics[width=.9\textwidth]{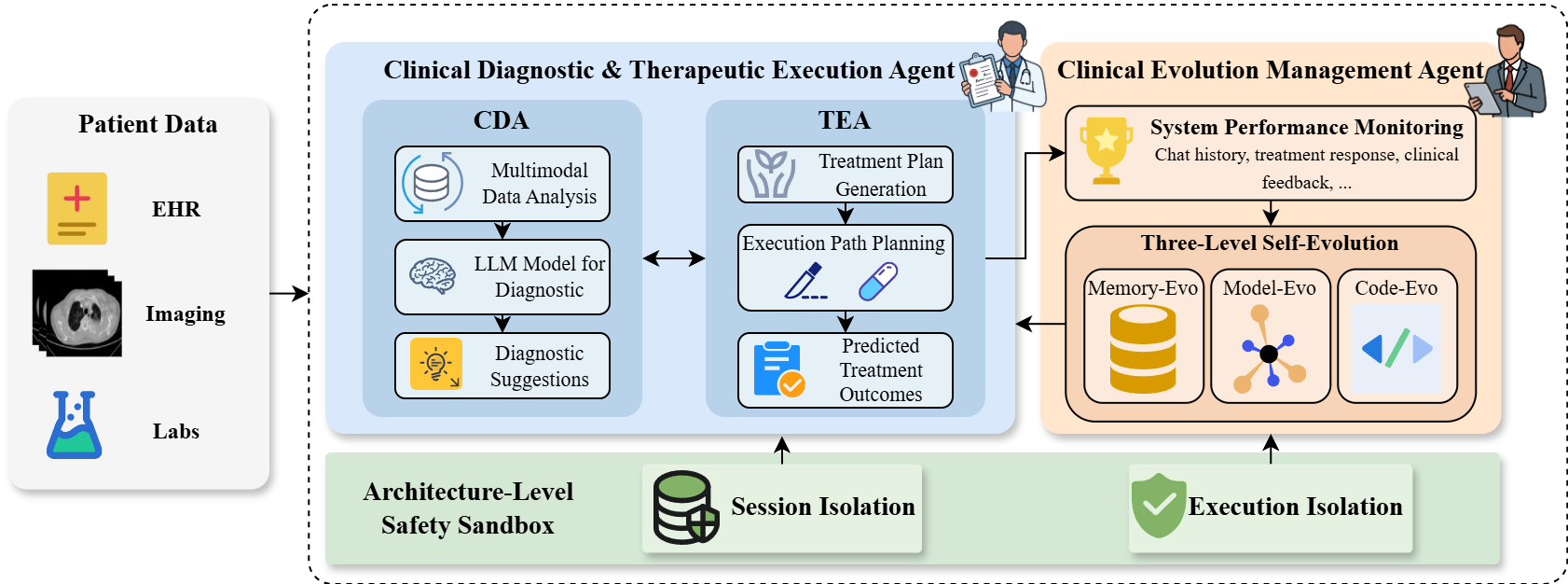}
	\caption{System architecture of the VIBEMed framework, consisting of the multi-agent collaborative framework, three-level self-evolution mechanism and an architecture-level safety sandbox.}
	\label{overview}
\end{figure*}

VIBEMed is an end-to-end multi-agent system designed to provide precision clinical decision support while ensuring safety and enabling autonomous improvement through interactions. Built on the paradigm of vibe medicine, it represents a fundamental transition from static, predefined pipelines to a dynamic, self-improving framework. The overall architecture of VIBEMed is presented in Figure \ref{overview}.

To support complex clinical workflows and real-world deployment requirements, VIBEMed is established based on three key mechanisms. First, a multi-agent collaborative framework decomposes the clinical decision-making process into specialized roles, enabling structured reasoning from diagnosis to treatment planning while maintaining coordination across stages. Second, a three-level self-evolution mechanism supports continuous improvement at the memory, model, and code levels, allowing the system to incorporate new data, refine its capabilities, and extend functionality over time. Third, an architecture-level safety sandbox enforces system reliability through execution and data isolation, ensuring that system updates and patient interactions remain controlled and traceable.

\subsection{Multi-agent collaborative framework}
\begin{figure*}[!htbp]
	\centering
	\includegraphics[width=.9\textwidth]{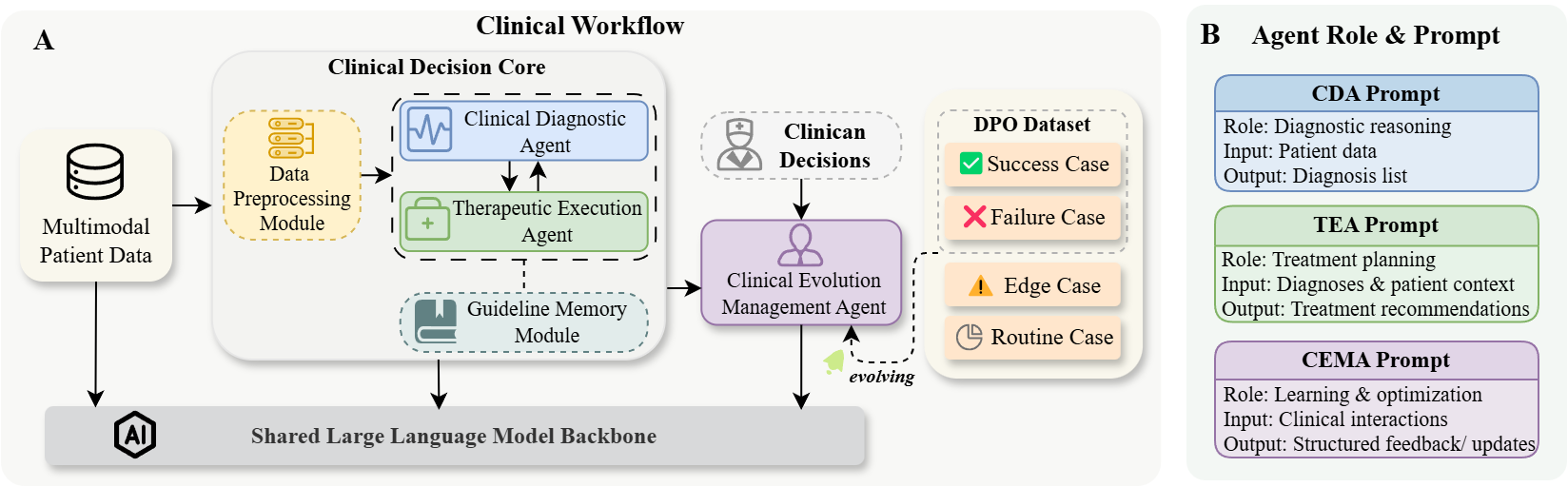}
	\caption{Overview of the multi-agent collaborative framework. \textbf{A:} Clinical workflow showing three agents (CDA, TEA, CEMA) powered by a shared backbone LLM with role-specific system prompts. \textbf{B:} Example for each agent showing role definitions with corresponding input \& output.}
	\label{3agents}
\end{figure*}

VIBEMed employs a multi-agent collaboration framework designed to simulate real-world clinical workflows through specialized role assignments and coordinated decision-making. It addresses the fundamental limitations of single-model architectures that are hard to excel in complex clinical tasks, which often require robust performance from diagnostic reasoning to treatment planning. Rather than relying on a single general-purpose model, VIBEMed decomposes the clinical decision-making process into three distinct roles, each handled by a specialized agent. As illustrated in Figure \ref{3agents}, all three agents share the same foundational large language model but are differentiated through role-specific system prompts that guide their behavior and output format. This design strikes a balance between specialization and simplicity, avoiding the computational overhead of maintaining multiple independent models while ensuring that each agent develops expertise in its designated domain. The Clinical Diagnostic Agent, Therapeutic Execution Agent, and Clinical Evolution Manager Agent operate collaboratively in a pipeline that transforms multimodal patient data into personalized treatment recommendations while continuously learning from clinical experience.

\textbf{Clinical Diagnostic Agent:} 
CDA serves as the entry point of the clinical decision-making pipeline, generating comprehensive diagnoses based on patient information presented in natural language. When it receives the patient analysis request through a user prompt, it automatically identifies relevant patient context and retrieves corresponding data from electronic health record\chdeleted{ systems} \chreplaced{and lab information systems}{without manual preprocessing steps}. \chadded{Medical imaging reports are incorporated in CDA instead of directly processing raw imaging data, since existing multimodal LLMs remain limited in accurately interpreting features in raw images for complex clinical decisions \cite{tavakoli2025generative}.} \chdeleted{This input includes chief complaints, past medical history, laboratory results, imaging reports, and other structured patient records.} \chadded{To enable effective diagnosis and treatment planning, an automated data preprocessing module is embedded within CDA to transform raw clinical text into structured format. In terms of radiology reports, several fields are extracted, including exam type, modality and sites, as well as diagnosis findings and conclusion, which facilitates effective integration of image-based information with other patient data such as chief complaints, past medical history, and laboratory results. These inputs are processed using predefined functions for de-identification and cleaning, together with LLM to standardize long content into key clinical fields.}

The agent processes \chreplaced{these structured inputs}{this input} and produces a ranked list of potential diagnoses, along with supporting evidence, suggested follow-up investigations to refine diagnostic certainty, and an assessment of clinical urgency. The confidence levels are determined based on the completeness of clinical information, the typicality of presentation, and the strength of evidence, rather than precise probability estimates that may convey false precision. The CDA is designed to account for uncertainty by considering multiple diagnostic possibilities and flagging cases with incomplete or atypical findings. This helps reduce errors caused by committing to a diagnosis too early and provides the downstream TEA with greater flexibility for treatment planning. \chdeleted{The structured output format supports seamless integration with subsequent stages while remaining interpretable to clinicians.}

\textbf{Therapeutic Execution Agent:} 
TEA receives diagnostic output from the CDA and translates it into actionable treatment recommendations tailored to individual patient. The TEA includes a dedicated guideline memory module that maintains structured clinical guidelines and continuously updated knowledge sources. During treatment planning, relevant guidance is retrieved from this module and combined with patient-specific conditions, ensuring that recommendations are grounded in medical evidence while remaining adaptable to the clinical context. The TEA also incorporates a three-tier safety constraint system that operates automatically before any recommendation is presented. The first tier represents absolute contraindications that block unsafe options, such as preventing bevacizumab in patients with recent hemoptysis due to bleeding risk. The second tier generates warnings for potential risks or out-of-range doses that require clinician acknowledgment but allow continuation. The third tier provides informational notices for off-label use and other low-risk situations, making sure that clinicians are aware of these choices. This design maintains safety while allowing flexibility in clinical decisions, with any deviations from standard practice clearly documented.

\chadded{To enable continuous incorporation of up-to-date medical knowledge, TEA adopts a hybrid retrieval and validation mechanism. Leveraging the tool-use capability of the agent system, TEA dynamically retrieves information from authoritative clinical guidelines, recent peer-reviewed literature, and curated medical knowledge bases. These sources are further transformed into structured representations through a RAG pipeline to support efficient semantic retrieval during downstream reasoning. The system then evaluates newly retrieved knowledge based on recency, source credibility, and level of evidence, and integrates qualified content into the Guideline Memory Module. To ensure stability and traceability, long-term memory is updated in a modular manner with different update frequencies based on the source type. The authoritative clinical guidelines are refreshed every six months since they are typically revised annually, while recent literature and medical knowledge bases are updated on a monthly basis.}
\chadded{To maintain the reliability of stored knowledge over time, TEA further incorporates an outdated knowledge management mechanism. When knowledge is integrated into the Guideline Memory Module, each entry is labeled with its publication date and last update time to ensure traceability. The system periodically re-evaluates stored knowledge against newly retrieved evidence; if substantial updates are identified or newer evidence demonstrates improved clinical outcomes, the corresponding older entries are assigned lower priority or marked as deprecated. Outdated knowledge is not immediately removed but compressed and summarized in an archived form to support reproducibility.}

\chadded{During treatment decision-making, TEA further incorporates a conflict-aware strategy to handle inconsistencies across clinical guidelines. When conflicts occur between clinical guidelines, TEA prioritizes recommendations based on guideline authority, publication time, and supporting evidence. In addition, considering the regional variability of clinical practice, the system gives preference to guidelines that are more consistent with the local healthcare context \cite{wang2020variations}, such as those published by the target regions. When conflicts cannot be fully resolved, TEA explicitly flags the uncertainty with reasons and relevant inconsistencies, allowing clinicians to make the final decision.
}

\textbf{Clinical Evolution Management Agent:} 
CEMA represents the learning component of the framework, responsible for extracting experience from completed interactions and converting it into structured knowledge to improve future performance. It receives comprehensive encounter records, including the dialogue history, diagnostic hypotheses generated by the CDA, treatment recommendations from the TEA, actions (adopted/modified/rejected) taken by clinicians, as well as follow-up outcomes such as treatment response, adverse events, and survival status. Based on this information, the CEMA classifies each case into four categories according to its learning value. Success cases are characterized by high clinician adoption rates (>80\%) and favorable patient outcomes, yielding positive patterns for reinforcement. \chadded{This threshold reflects a level of agreement that substantially exceeds typical clinical decision support systems adoption rates reported in real-world settings \cite{newton2025systematic}.}
\chadded{Based on clinical feedback,} failure cases are associated with high clinician rejection rates (>50\%) or adverse events noticed by the system, providing lessons for error avoidance. Edge cases encompass rare conditions, special populations, or situations involving conflicting guidelines, which can help expand system capability boundaries. Routine cases that do not meet these criteria are summarized and stored, but do not trigger evolution processes.

\chadded{As illustrated in Figure \ref{3agents},} for success and failure cases, the CEMA generates structured reflection datasets in a contrastive format suitable for DPO training,\chdeleted{as illustrated in Figure \ref{3agents}.} \chadded{where input contains the patient information and output corresponds to the accepted or rejected treatment details.} These datasets capture not only system recommendations and clinician decisions, but also the reasoning behind the differences, enabling the model to learn from both correct and incorrect decisions. \chadded{The reflection dataset is maintained dynamically within CEMA, containing approximately 40k samples including both success and failure cases in roughly equal proportion, which is updated weekly with newly collected interaction records based on a FIFO manner. Each sample is associated with a timestamp, allowing the system to prioritize more recent cases during training.}

\subsection{Three-level self-evolution mechanism}
\begin{figure*}[!htbp]
	\centering
	\includegraphics[width=.9\textwidth]{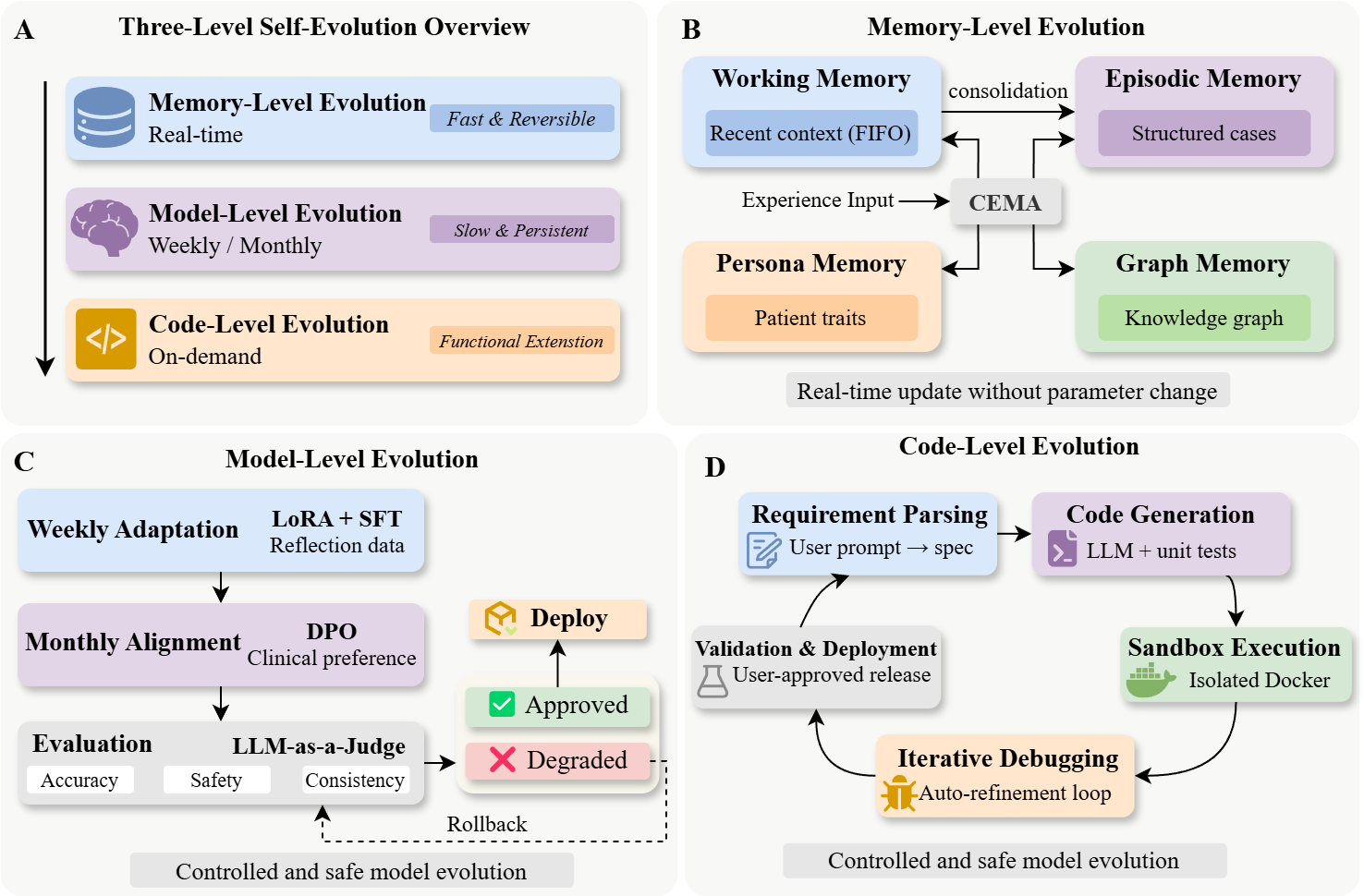}
	\caption{Details of the three-level self-evolution mechanism. \textbf{A:} Conceptual overview of the hierarchical structure encompassing memory, model, and code evolutions. \textbf{B:} Memory-level evolution showing real-time updates through a multi-dimensional storage without parameter change. \textbf{C:} Model-level evolution featuring periodic adaptation using Supervised Fine-Tuning (SFT) and Direct Preference Optimization (DPO), governed by the ``LLM-as-a-judge'' safety evaluation loop. \textbf{D:} Code-level evolution workflow demonstrating the on-demand functional extension process.}
	\label{evolve}
\end{figure*}

VIBEMed implements a three-level self-evolution mechanism that enables continuous improvement across memory, model, and code without requiring frequent intervention or periodic retraining. This design is inspired by human learning, where immediate experience is first held in working memory, then consolidated into long-term memory through reflection and repetition, and eventually internalized as intuitive responses through practice. As shown in Figure \ref{evolve}, the system performs memory-level evolution for rapid adaptation through experience storage and retrieval, model-level evolution for sustained capability improvement through parameter updates, and code-level evolution for functional extension through automated code modification.
Each layer operates at a different timescale and computational cost, balancing responsiveness with stability in clinical settings.

\textbf{Memory-Level Evolution}: 
At the memory level, VIBEMed leverages a hierarchical memory architecture that enables efficient storage and retrieval of clinical experience. Working memory maintains recent conversational context within a bounded FIFO window, allowing agents to preserve coherence during multi-turn interactions without accessing long-term storage. When capacity is reached, earlier histories are consolidated into episodic memory, which stores structured records organized by condition, treatment, and outcome. Persona memory contains stable patient characteristics including comorbidities, allergies, and treatment preferences, enabling personalized recommendations without requiring re-extraction from raw records. Graph memory implements a knowledge graph connecting diseases, treatments, outcomes, and patient attributes through typed edges with weights, supporting relation-aware retrieval for causal and temporal reasoning beyond simple semantic matching. In parallel, the guideline memory module in TEA is maintained and updated, ensuring that treatment decisions remain aligned with current clinical standards. The CEMA determines the distribution of newly extracted experience across these memory components based on its characteristics and expected reuse, balancing specificity for accurate retrieval and generalization for broad applicability. Memory-level evolution operates in real-time, with updates applied after each completed encounter, offering rapid adaptation without requiring model retraining. Since memory updates do not modify model parameters, they can be rolled back immediately if issues are detected, providing a safe mechanism for continuous learning.

\textbf{Model-Level Evolution}: 
At the model level, VIBEMed adopts a staged training strategy that balances rapid adaptation with long-term stability. \chadded{Considering the trade-offs between training stability and computational efficiency based on literature suggestions \cite{liu2025generalist},} the weekly adaptation phase applies Low-Rank Adaptation (LoRA) with SFT on \chadded{1k-3k success cases from} reflection dataset\chdeleted{ constructed from high-quality cases}, enabling efficient updates with minimal parameter changes. The monthly training phase leverages DPO on \chreplaced{5k-10k paired success and failure cases}{contrastive data} derived from clinician decisions, aligning model behavior with real-world clinical preferences.

To ensure reliable deployment, model updates are introduced in a controlled and progressive manner. Newly trained models are first evaluated alongside the current production model on the same inputs, with their outputs assessed using automated metrics and LLM-as-a-judge \cite{gu2024survey} across dimensions such as clinical accuracy, safety, and consistency with clinician behavior. If the new model demonstrates stable or improved performance, it would be introduced into production; otherwise, the system can immediately revert to the previous version without interrupting service if any degradation is observed. This design ensures that model evolution remains both effective and safe, allowing continuous improvement while maintaining stable clinical performance.

\textbf{Code-Level Evolution}: 
At the code level, VIBEMed implements an automated code evolution mechanism that enables capability extension through code generation, validation, and deployment. This layer complements model-level updates by supporting agent interaction optimization and functional updates. In particular, interaction optimization is achieved by analyzing execution traces and inter-agent message flows, and translating these signals into modifications of prompt templates, tool invocation strategies, and agent routing logic. Additionally, when unsupported functionality is requested by the user, the system initiates a structured workflow by formalizing natural language input into a specification. Implementation code and corresponding unit tests are then generated using the backbone LLM, with optional support of alternative code generation models. The generated code is executed in an isolated Docker sandbox with restricted permissions, where unit tests are run to verify correctness. If tests fail, the system iteratively refines the code until predefined criteria are met. The validated functionality is then presented in the test environment, and deployment to production proceeds only after user confirmation. 

\subsection{Architecture-level safety sandbox}
VIBEMed implements an architecture-level safety sandbox to guarantee safe operation through isolated execution environments and context-based data separation. Instead of relying on agent behavior or manual oversight, safety is enforced through system constraints that restrict unauthorized actions and ensure traceability. The design consists of two components. Execution isolation separates development, validation, and production stages, preventing unverified code from reaching clinical use. In the meantime, memory isolation enforces strict separation across patient encounters through indexed access control applied consistently across all memory tiers. These mechanisms allow the system to operate autonomously while maintaining the reliability required for clinical settings.

\textbf{Execution Environments Isolation:}
VIBEMed adopts a multi-environment execution architecture to isolate system modification from clinical deployment. The framework maintains separate development, validation, and production stages. Newly generated or modified code is first executed within a sandbox development environment with restricted permissions, where automated validation is performed. Verified components are then promoted to a validation stage for scenario-based evaluation, and subsequently to production after passing all checks. For each new functionality, the system generates a structured disclosure report prior to activation. The report summarizes data access scope, affected clinical workflows, communication interfaces, and potential operational risks identified through static analysis and predefined risk patterns. Activation requires explicit user acknowledgment of this report. System activities are recorded through an immutable audit mechanism. Key events, including tool invocations, data access, and recommendation generation, are logged to an append-only store with cryptographic guarantees. Logging is implemented at the system level, independent of agent execution.

\textbf{Session-Level Memory Isolation:}
Due to the sensitivity of patient data, VIBEMed enforces a strict isolation policy through a session-indexed memory mechanism. Each clinical session is assigned a unique identifier and allocated an independent memory space with an associated access control list. All data retrieved during the session, including demographics, medical history, laboratory results, and imaging findings, are tagged with this identifier and stored within the corresponding scope. Access control policies ensure that agents can only read or write data associated with the current context, preventing cross-patient access. This mechanism is applied consistently across all memory tiers. Working memory remains session-scoped and is cleared upon completion. Episodic memory stores encounter records with explicit identifiers and requires matching credentials for retrieval. Persona memory is linked to persistent patient identifiers, with access gated by session-level authorization. Graph memory enforces relation-level tagging to prevent context-specific associations from being incorporated into the global graph without validation. Access control is enforced through database row-level security together with application-side checks, providing layered protection against unintended or unauthorized data access.

\section{Results}

We evaluated VIBEMed from both quantitative performance and system-level validation. \chadded{We conducted a dedicated evaluation of the multi-agent collaboration framework to assess its ability to detect and mitigate the propagation of upstream errors into downstream clinical decisions.}
\chreplaced{In addition}{First}, we assessed \chdeleted{the effectiveness of }the self-evolution mechanism through benchmark experiments, focusing on its impact on medical reasoning and decision-making capabilities. \chdeleted{Second, }We \chadded{also} examined the system's practical usability through an end-to-end interface demonstration in complex clinical scenarios. In conclusion, these results provide evidence that VIBEMed can both improve model performance through continuous evolution and support reliable clinical decision-making in real-world settings.

\subsection{Evaluation of the multi-agent framework}
\chadded{To assess the performance and reliability of the proposed multi-agent collaboration framework, we constructed a dedicated evaluation dataset comprising 30 complex clinical cases. These cases cover three categories of key clinical challenges, including 11 dynamic progression cases, 9 multimorbidity cases, and 10 diagnostic ambiguity cases. An upstream clinical error will be embedded within each case, which would propagate to downstream treatment decisions and potentially lead to unsafe outcomes if not identified. The dataset was generated using advanced Claude Opus 4.6 under clinical guidelines constraints. Simulated cases were used instead of real medical records because this study requires the embedding of known and traceable upstream errors in each case as evaluation benchmarks, which is difficult to achieve with real retrospective data. Similar approaches have been widely adopted in the evaluation of AI medical systems and have demonstrated high efficiency and effectiveness across multiple clinical domains \cite{jarrassier2026can}.}

\chadded{Four pipelines were leveraged to assess each component independently and systematically. Pipeline A is the direct generation pipeline, where raw cases are processed by a general-purpose LLM without any agent structure. Pipeline B and C only utilizes CDA and TEA, respectively. Pipeline D is the complete two-stage cascade process consisting of the CDA followed by the TEA. All ablation experiments were performed using DeepSeek-V3.2, with a generation temperature set to 0.30. An independent LLM, Qwen3.6 Plus, was employed for the output evaluation, with a scoring temperature of 0.20 to ensure consistency. These state-of-the-art models were selected due to their large parameter scales and strong reasoning capabilities, making them suitable for reliable generation and assessment. Each pipelines was evaluated based on eight different clinical quality metrics, including diagnostic accuracy, upper level error detection, appropriateness of treatment, safety and contraindication Identifying, prevention of error propagation, clinical completeness, transparency of reasoning and dynamic adaptability. The metrics was scored on a 1-5 point scale with the aggregate being a total of 40 points.}

\chadded{As shown in Table \ref{tab:avg-score}, Pipeline D achieved the highest mean score of 39.20 and lowest standard deviation of 1.42 across the 30 cases, indicating that the multi-agent framework enhances overall performance and greatly reduces variability in output quality. Besides, Pipeline D had a 2.23 point improvement compared to Pipeline B, demonstrating that TEA can actively identify and provides compensation for missed upstream diagnostic outputs after receiving the results from CDA. This will also help to further complete missing data, quantify risks, and refine treatment plans. Pipeline D also had the largest difference of 7.67 compared to Pipeline C, since TEA is expected to utilize the structured diagnostic information provided by the CDA when it develops its treatment plans, which proves the importance of upstream diagnostic support. In terms of the single model generating process used in Pipeline A, Pipeline D had a 4.07 point increase in performance and a lower standard deviation, showing the effectiveness of the multi-agent system. Based on the empirical results, the proposed multi-agent collaboration framework is capable of detecting and reducing the transmission of upstream errors into treatment decisions, demonstrating safer and more reliable performance than single model approaches in more complicated clinical settings.}

\begin{table}[!htbp]
\centering
\caption{Evaluation scores of the four pipelines.}
\label{tab:avg-score}
\begin{tabular}{c|llll}
\hline\hline
\textbf{Experiments} & \multicolumn{1}{c}{\textbf{Avg}} & \textbf{SD} & \multicolumn{1}{c}{\textbf{Min}} & \multicolumn{1}{c}{\textbf{Max}} \\ \hline
\textbf{Pipeline A: Direct LLM} & 35.13 & 3.92 & 24 & 40 \\
\textbf{Pipeline B: CDA only} & 36.97 & 1.61 & 32 & 40 \\
\textbf{Pipeline C: TEA only} & 31.53 & 4.75 & 24 & 39 \\
\textbf{Pipeline D: CDA + TEA} & 39.20 & 1.42 & 35 & 40 \\ \hline\hline
\end{tabular}
\end{table}

\subsection{Effectiveness of self-evolution mechanism}
To validate the effectiveness of the proposed self-evolution mechanism, we compared the performance of the base model against the evolved model after applying the evolution strategy. DeepSeek-Distill-Qwen2.5-1.5B was selected as the backbone architecture to optimize computational cost and experiment efficiency. Previous studies demonstrate that scaling laws allow performance gains observed in smaller models to reliably predict similar improvements in larger models \cite{kaplan2020scaling}, supporting its use as a proxy for assessing the impact of the evolution mechanism.

\chadded{We adopted SFT with LoRA on 8$\times$H20 GPUs for 3 epochs to enable efficient training, leveraging a learning rate of 5e-4 and a cosine scheduler. The per-device training batch size was set to 1 and bfloat16 precision was employed for improved training stability. Regarding the LoRA configuration, a rank scaling factor of $\alpha$ = 32 and a dropout rate = 0.05 was used. The training dataset consists of approximately 4k medical treatment samples, along with an additional 10k distilled general-domain samples to resolve the catastrophic forgetting problem while preserving the model's general reasoning capability during domain adaptation.}

The evaluation was conducted on the MedBench complex medical reasoning dataset developed by OpenCompass \cite{liu2024medbench, ding2025medbench}. \chadded{By leveraging three state-of-the-art LLMs, Intern-S1, Qwen3-235B-A22B, and DeepSeek-R1, this ``LLM-as-a-judge'' framework can demonstrate reliable and comparable performance to human experts under well-defined validation criteria \cite{zhou2025automating}. They assess clinical outputs across various predefined scoring criteria, covering factual correctness, clinical safety, reasoning consistency, and completeness of key medical information.} We selected several clinically relevant \chreplaced{benchmarks}{subsets} including clinical reasoning (CMB-Clin-extended), treatment planning (MedTreat), outcome prediction (MedOutcome), risk assessment (MedAnalysis), diagnosis (MedDiag), differential diagnosis (MedDiffer), and personalized health management (MedPHM) \chadded{to cover core clinical scenarios. In practice, LLM-based evaluation is primarily used for efficient large-scale model screening, while final deployment decisions can be further supported by human expert review to ensure clinical reliability and safety.}

\begin{table}[!htbp]
\centering
\caption{Performance comparison between the base model and the evolved model on MedBench clinical reasoning tasks.}
\label{tab:evovle}
\begin{tabular}{l|l|l}
\hline\hline
 & \multicolumn{1}{c|}{\textbf{Base Model}} & \multicolumn{1}{c}{\textbf{Evolved Model}} \\ \hline
\textbf{CMB-Clin-extended} & 22.5 & \textbf{23.4} \\
\textbf{MedTreat} & 21.6 & \textbf{25.5} \\
\textbf{MedOutcome} & 18.0 & \textbf{18.7} \\
\textbf{MedAnalysis} & 38.8 & 35.8 \\
\textbf{MedDiag} & 26.7 & 24.7 \\
\textbf{MedDiffer} & 9.3 & \textbf{13.0} \\
\textbf{MedPHM} & 30.8 & \textbf{43.2} \\ \hline\hline
\end{tabular}
\end{table}

As shown in Table \ref{tab:evovle}, the evolved model demonstrates consistent improvements across several key dimensions. Substantial gains were observed in MedTreat (+3.9) and MedPHM (+12.4), indicating a marked improvement in personalized treatment planning and long-term patient management. Performance on MedDiffer (+3.7) also improved, suggesting enhanced capability in differential diagnosis, which is critical for handling complex or ambiguous clinical cases. In addition, moderate improvements in CMB-Clin-extended and MedOutcome support the effectiveness of the evolution mechanism in real-world clinical reasoning and outcome prediction tasks. However, performance declines were noticed in MedAnalysis (-3.0) and MedDiag (-2.0). These tasks rely more heavily on precise numerical reasoning or standardized diagnostic patterns, which are less likely to be influenced by experience-driven evolution. This suggests that the current evolution strategy primarily benefits higher-level clinical reasoning and decision-making, rather than tightly structured or calculation-heavy tasks.

Overall, the results indicate that the self-evolution mechanism driven by CEMA can effectively improve the model's ability to handle complex clinical decision-making scenarios. The gains are particularly evident in tasks requiring integrated reasoning, personalization, and long-term planning, which are central to real-world medical practice. These findings support the role of CEMA in enabling continuous, experience-driven improvement within the VIBEMed framework.

\subsection{System visualization and validation}
\begin{figure*}[!htbp]
	\centering
	\includegraphics[width=.9\textwidth]{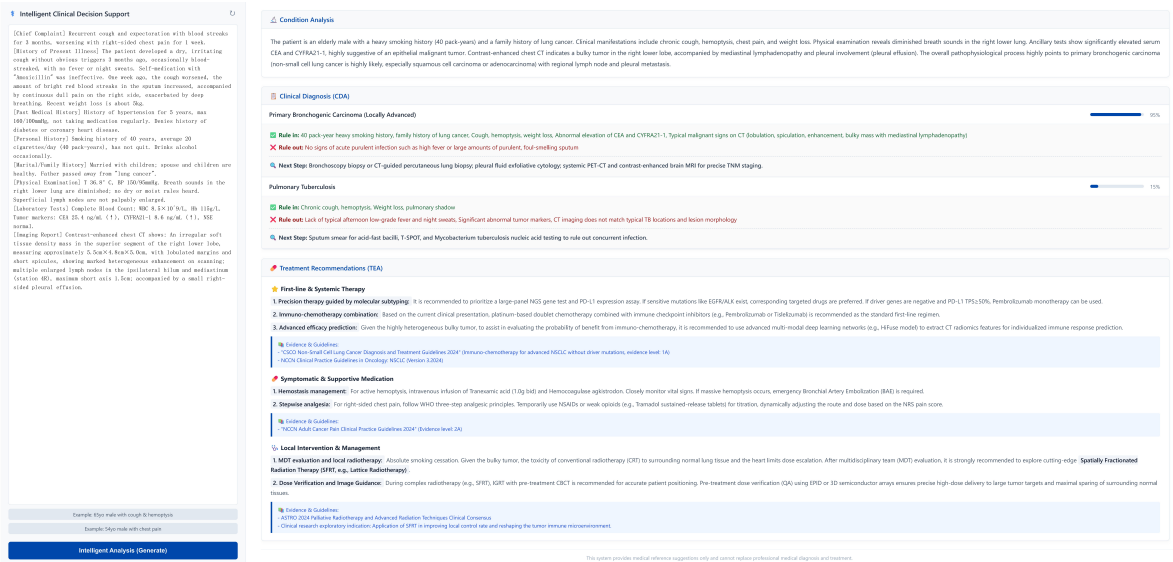}
	\caption{Functional interface of the VIBEMed system. The left panel displays the input of multimodal patient data. The right panel presents the agent-collaborative decision outputs from CDA and TEA.}
	\label{platform}
\end{figure*}

To evaluate the clinical feasibility and operational robustness of the VIBEMed framework, we validated its end-to-end performance in highly complex clinical scenarios, particularly an advanced-stage cancer case with severe complications. Figure \ref{platform} visualizes the system platform, which provides a clear demonstration of how the system transforms unstructured, multimodal patient data into a standardized clinical decision-making workflow, proving its practical value in real-world medical settings.

During the diagnostic phase, the CDA demonstrated exceptional data parsing and uncertainty management capabilities. By extracting relevant features from the multimodal inputs including chief complaints, physical examinations, laboratory tests, and imaging reports, it automatically generated a structured differential diagnosis list. \chadded{In particular, the imaging report provides critical diagnostic evidence for this case. The system identifies key radiological features, including a lobulated mass with spiculated margins in the right lower lobe, heterogeneous enhancement, and associated mediastinal lymphadenopathy and pleural effusion, supporting a high likelihood of malignancy.} Instead of producing a single deterministic output, the CDA assigned confidence levels to each candidate diagnosis and explicitly linked them to supporting and excluding evidence. It also suggested follow-up examinations to refine diagnostic certainty, such as biopsy or additional imaging. This design promotes transparent reasoning and helps avoid errors associated with making a diagnosis before sufficient evidence is available.

In the treatment planning phase, the TEA generated patient-specific treatment strategies by integrating diagnostic results with clinical guidelines. The guideline memory module allowed the system to retrieve and apply up-to-date evidence while adapting to individual patient conditions. In addition, safety constraints were enforced throughout this process as part of the system design. For high-risk or non-standard scenarios, the system flagged potential concerns or recommended multidisciplinary team evaluation when necessary, ensuring the reliability of the clinical decisions.

To facilitate practical deployment, VIBEMed is designed to be compatible with existing hospital infrastructure. When generating advanced radiotherapy strategies, such as space-resolved radiotherapy (SFRT) for large-volume tumors, the system aligns with standard clinical physics quality control workflows, including cone-beam CT (CBCT)-based target localization verification and electronic portal imaging device (EPID)-based beam geometry verification. This compatibility allows the system to be integrated into routine clinical practice without requiring substantial changes to existing hardware or workflows, indicating its potential for real-world adoption.

\section{Conclusions}

Recent advances in large language models have demonstrated strong potential in medical applications, particularly in structured or knowledge-based tasks. However, their real-world clinical utility remains limited due to insufficient grounding, lack of workflow integration, and inadequate safety control \cite{chen2026llm}. Therefore, we propose VIBEMed, an end-to-end clinical decision support system that integrates a multi-agent collaborative framework, a three-level self-evolution mechanism, and an architecture-level safety sandbox to address this limitation. By coordinating three specialized agents, CDA, TEA, and CEMA, and enabling adaptive updates across memory, model, and code, VIBEMed supports experience-driven improvement beyond static model capabilities.

The three-agent framework reflects key aspects of real-world clinical workflows and produces practical, context-aware decisions that can be refined through interaction. The empirical results suggest that incorporating structured evolution mechanisms can improve complex clinical reasoning performance without compromising system stability. At the same time, the safety sandbox provides system-level constraints that regulate execution and data access, which is critical for maintaining reliable operation when handling sensitive patient information. This combination addresses a central gap in current medical AI systems, which tends to show strong performance in isolated tasks but lack the ability to evolve safely within real-world settings.

Nevertheless, there still exist several limitations to be acknowledged. First, the experiments were conducted using a 1.5B-parameter backbone model for computational efficiency. While this allows for controlled evaluation, the small model size limits the system’s ability to process complex information, and further validation on larger models is necessary to fully assess its potential. Second, the current evaluation is primarily based on benchmark datasets, and additional validation through prospective clinical studies across multiple institutions is required to establish clinical reliability and interpretability. \chadded{In addition, the three-level evolution framework is not fully evaluated in terms of the individual contribution of each component. Although several safety mechanisms are incorporated, their effectiveness has not been systematically assessed through dedicated experiments.} In the future, we would extend the framework to larger backbone models for evaluation in scalability and performance with higher-capacity settings. Besides, it is advised to conduct multi-center studies, including retrospective and prospective validation, to assess robustness and generalizability in real clinical environments. \chadded{Hierarchical ablation studies will be performed to quantify the necessity of different evolution levels, along with more experiments to fully validate the proposed safety mechanisms.}

In summary, this work presents a practical step toward clinical AI systems that can improve through experience while remaining controllable and reliable. VIBEMed provides a novel framework for exploring this direction and contributes to the development of decision support systems that better align with the dynamic nature of clinical practice.


\section*{Declaration of generative AI and AI-assisted technologies in the manuscript preparation process}

During the preparation of this work the author(s) used OpenClaw and ChatGPT in order to conduct literature research, refine the linguistic quality of the manuscript, and evaluate the logical consistency of the arguments. After using this tool/service, the author(s) reviewed and edited the content as needed and take(s) full responsibility for the content of the published article.

\section*{Acknowledgments}
This work was supported by the National Foreign Experts Program (S20250235), Chronic Disease Management Research Program of the Center for Capacity Building and Continuing Education, National Health Commission (GWJJMB202510022202), Hebei Provincial "Yanzhao Golden Terrace Talent Recruitment Program" - Top-Tier Talent Program (HY2025050008), The Innovation \& Development Medical Cooperation Program of Hengrui-Hebei (HR202502085), Hebei Provincial Health Commission Medical Research and Enterprise Joint Innovation Program (LH20250086), Hebei Provincial Health Commission Medical Research Program (20250011, 20260208, 20260195), Spark Scientific Research Program of The First Hospital of Hebei Medical University (XH202515).

\bibliographystyle{elsarticle-num-names}

\bibliography{main}

@article{thirunavukarasu2023large,
  title={Large language models in medicine},
  author={Thirunavukarasu, Arun James and Ting, Darren Shu Jeng and Elangovan, Kabilan and Gutierrez, Laura and Tan, Ting Fang and Ting, Daniel Shu Wei},
  journal={Nature medicine},
  volume={29},
  number={8},
  pages={1930--1940},
  year={2023},
  publisher={Nature Publishing Group US New York}
}

@article{shen2023chatgpt,
  title={ChatGPT and other large language models are double-edged swords},
  author={Shen, Yiqiu and Heacock, Laura and Elias, Jonathan and Hentel, Keith D and Reig, Beatriu and Shih, George and Moy, Linda},
  journal={Radiology},
  volume={307},
  number={2},
  pages={e230163},
  year={2023},
  publisher={Radiological Society of North America}
}

@article{li2025care,
  title={CARE-AD: a multi-agent large language model framework for Alzheimer’s disease prediction using longitudinal clinical notes},
  author={Li, Rumeng and Wang, Xun and Berlowitz, Dan and Mez, Jesse and Lin, Honghuang and Yu, Hong},
  journal={npj Digital Medicine},
  volume={8},
  number={1},
  pages={541},
  year={2025},
  publisher={Nature Publishing Group UK London}
}

@article{peng2024integration,
  title={Integration of multi-source medical data for medical diagnosis question answering},
  author={Peng, Qi and Cai, Yi and Liu, Jiankun and Zou, Quan and Chen, Xing and Zhong, Zheng and Wang, Zefeng and Xie, Jiayuan and Li, Qing},
  journal={IEEE Transactions on Medical Imaging},
  volume={44},
  number={3},
  pages={1373--1385},
  year={2024},
  publisher={IEEE}
}

@article{vaswani2017attention,
  title={Attention is all you need},
  author={Vaswani, Ashish and Shazeer, Noam and Parmar, Niki and Uszkoreit, Jakob and Jones, Llion and Gomez, Aidan N and Kaiser, {\L}ukasz and Polosukhin, Illia},
  journal={Advances in neural information processing systems},
  volume={30},
  year={2017}
}

@article{singhal2023large,
  title={Large language models encode clinical knowledge},
  author={Singhal, Karan and Azizi, Shekoofeh and Tu, Tao and Mahdavi, S Sara and Wei, Jason and Chung, Hyung Won and Scales, Nathan and Tanwani, Ajay and Cole-Lewis, Heather and Pfohl, Stephen and others},
  journal={Nature},
  volume={620},
  number={7972},
  pages={172--180},
  year={2023},
  publisher={Nature Publishing Group UK London}
}

@article{zheng2025large,
  title={Large language models for medicine: a survey},
  author={Zheng, Yanxin and Gan, Wensheng and Chen, Zefeng and Qi, Zhenlian and Liang, Qian and Yu, Philip S},
  journal={International Journal of Machine Learning and Cybernetics},
  volume={16},
  number={2},
  pages={1015--1040},
  year={2025},
  publisher={Springer}
}

@article{hou2024large,
  title={Large language models for software engineering: A systematic literature review},
  author={Hou, Xinyi and Zhao, Yanjie and Liu, Yue and Yang, Zhou and Wang, Kailong and Li, Li and Luo, Xiapu and Lo, David and Grundy, John and Wang, Haoyu},
  journal={ACM Transactions on Software Engineering and Methodology},
  volume={33},
  number={8},
  pages={1--79},
  year={2024},
  publisher={ACM New York, NY}
}

@article{kelly2019key,
  title={Key challenges for delivering clinical impact with artificial intelligence},
  author={Kelly, Christopher J and Karthikesalingam, Alan and Suleyman, Mustafa and Corrado, Greg and King, Dominic},
  journal={BMC medicine},
  volume={17},
  number={1},
  pages={195},
  year={2019},
  publisher={Springer}
}

@article{shi2025continual,
  title={Continual learning of large language models: A comprehensive survey},
  author={Shi, Haizhou and Xu, Zihao and Wang, Hengyi and Qin, Weiyi and Wang, Wenyuan and Wang, Yibin and Wang, Zifeng and Ebrahimi, Sayna and Wang, Hao},
  journal={ACM Computing Surveys},
  volume={58},
  number={5},
  pages={1--42},
  year={2025},
  publisher={ACM New York, NY}
}

@article{lewis2020retrieval,
  title={Retrieval-augmented generation for knowledge-intensive nlp tasks},
  author={Lewis, Patrick and Perez, Ethan and Piktus, Aleksandra and Petroni, Fabio and Karpukhin, Vladimir and Goyal, Naman and K{\"u}ttler, Heinrich and Lewis, Mike and Yih, Wen-tau and Rockt{\"a}schel, Tim and others},
  journal={Advances in neural information processing systems},
  volume={33},
  pages={9459--9474},
  year={2020}
}

@article{peng2025graph,
  title={Graph retrieval-augmented generation: A survey},
  author={Peng, Boci and Zhu, Yun and Liu, Yongchao and Bo, Xiaohe and Shi, Haizhou and Hong, Chuntao and Zhang, Yan and Tang, Siliang},
  journal={ACM Transactions on Information Systems},
  volume={44},
  number={2},
  pages={1--52},
  year={2025},
  publisher={ACM New York, NY}
}

@article{zhong2025considerations,
  title={Considerations for patient privacy of Large Language Models in health care: scoping review},
  author={Zhong, Xiaoying and Li, Siyi and Chen, Zhao and Ge, Long and Yu, Dongdong and Wang, Shijia and You, Liangzhen and Shang, Hongcai},
  journal={Journal of Medical Internet Research},
  volume={27},
  pages={e76571},
  year={2025},
  publisher={JMIR Publications Toronto, Canada}
}

@article{haltaufderheide2024ethics,
  title={The ethics of ChatGPT in medicine and healthcare: a systematic review on Large Language Models (LLMs)},
  author={Haltaufderheide, Joschka and Ranisch, Robert},
  journal={NPJ digital medicine},
  volume={7},
  number={1},
  pages={183},
  year={2024},
  publisher={Nature Publishing Group UK London}
}

@article{chen2024integration,
  title={Integration of large language models and federated learning},
  author={Chen, Chaochao and Feng, Xiaohua and Li, Yuyuan and Lyu, Lingjuan and Zhou, Jun and Zheng, Xiaolin and Yin, Jianwei},
  journal={Patterns},
  volume={5},
  number={12},
  year={2024},
  publisher={Elsevier}
}

@article{lyu2026evoscientist,
  title={EvoScientist: Towards Multi-Agent Evolving AI Scientists for End-to-End Scientific Discovery},
  author={Lyu, Yougang and Zhang, Xi and Yi, Xinhao and Zhao, Yuyue and Guo, Shuyu and Hu, Wenxiang and Piotrowski, Jan and Kaliski, Jakub and Urbani, Jacopo and Meng, Zaiqiao and others},
  journal={arXiv preprint arXiv:2603.08127},
  year={2026}
}

@article{meske2025vibe,
  title={Vibe coding as a reconfiguration of intent mediation in software development: Definition, implications, and research agenda},
  author={Meske, Christian and Hermanns, Tobias and Von der Weiden, Esther and Loser, Kai-Uwe and Berger, Thorsten},
  journal={IEEE Access},
  volume={13},
  pages={213242--213259},
  year={2025},
  publisher={IEEE}
}

@article{fang2025comprehensive,
  title={A comprehensive survey of self-evolving ai agents: A new paradigm bridging foundation models and lifelong agentic systems},
  author={Fang, Jinyuan and Peng, Yanwen and Zhang, Xi and Wang, Yingxu and Yi, Xinhao and Zhang, Guibin and Xu, Yi and Wu, Bin and Liu, Siwei and Li, Zihao and others},
  journal={arXiv preprint arXiv:2508.07407},
  year={2025}
}

@article{gu2024survey,
  title={A survey on llm-as-a-judge},
  author={Gu, Jiawei and Jiang, Xuhui and Shi, Zhichao and Tan, Hexiang and Zhai, Xuehao and Xu, Chengjin and Li, Wei and Shen, Yinghan and Ma, Shengjie and Liu, Honghao and others},
  journal={The Innovation},
  year={2024},
  publisher={Elsevier}
}

@article{kaplan2020scaling,
  title={Scaling laws for neural language models},
  author={Kaplan, Jared and McCandlish, Sam and Henighan, Tom and Brown, Tom B and Chess, Benjamin and Child, Rewon and Gray, Scott and Radford, Alec and Wu, Jeffrey and Amodei, Dario},
  journal={arXiv preprint arXiv:2001.08361},
  year={2020}
}

@article{liu2024medbench,
  title={Medbench: A comprehensive, standardized, and reliable benchmarking system for evaluating chinese medical large language models},
  author={Liu, Mianxin and Hu, Weiguo and Ding, Jinru and Xu, Jie and Li, Xiaoyang and Zhu, Lifeng and Bai, Zhian and Shi, Xiaoming and Wang, Benyou and Song, Haitao and others},
  journal={Big Data Mining and Analytics},
  volume={7},
  number={4},
  pages={1116--1128},
  year={2024},
  publisher={TUP}
}

@article{chen2026llm,
  title={LLM-assisted systematic review of large language models in clinical medicine},
  author={Chen, Sully F and Alyakin, Anton and Seas, Andreas and Yang, Eunice and Choi, Joanne J and Lee, Jin Vivian and Chen, Amelia L and Warman, Pranav I and Bitolas, Rochelle T and Steele, Robert J and others},
  journal={Nature medicine},
  pages={1--8},
  year={2026},
  publisher={Nature Publishing Group US New York}
}

@article{wang2020variations,
  title={Variations in processes for guideline adaptation: a qualitative study of World Health Organization staff experiences in implementing guidelines},
  author={Wang, Zhicheng and Grundy, Quinn and Parker, Lisa and Bero, Lisa},
  journal={BMC Public Health},
  volume={20},
  number={1},
  pages={1758},
  year={2020},
  publisher={Springer}
}

@article{tavakoli2025generative,
  title={Generative AI and Foundation Models in Radiology: Applications, Opportunities, and Potential Challenges},
  author={Tavakoli, Neda and Shakeri, Zahra and Gowda, Vrushab and Samsel, Konrad and Bedayat, Arash and Ghasemiesfe, Ahmadreza and Bagci, Ulas and Hsiao, Albert and Leiner, Tim and Carr, James and others},
  journal={Radiology},
  volume={317},
  number={2},
  pages={e242961},
  year={2025},
  publisher={Radiological Society of North America}
}

@article{newton2025systematic,
  title={A systematic review of clinicians’ acceptance and use of clinical decision support systems over time},
  author={Newton, Nicki and Bamgboje-Ayodele, Adeola and Forsyth, Rowena and Tariq, Amina and Baysari, Melissa T},
  journal={npj Digital Medicine},
  volume={8},
  number={1},
  pages={309},
  year={2025},
  publisher={Nature Publishing Group UK London}
}

@article{liu2025generalist,
  title={A generalist medical language model for disease diagnosis assistance},
  author={Liu, Xiaohong and Liu, Hao and Yang, Guoxing and Jiang, Zeyu and Cui, Shuguang and Zhang, Zhaoze and Wang, Huan and Tao, Liyuan and Sun, Yongchang and Song, Zhu and others},
  journal={Nature medicine},
  volume={31},
  number={3},
  pages={932--942},
  year={2025},
  publisher={Nature Publishing Group US New York}
}

@article{ding2025medbench,
  title={MedBench v4: A Robust and Scalable Benchmark for Evaluating Chinese Medical Language Models, Multimodal Models, and Intelligent Agents},
  author={Ding, Jinru and Lu, Lu and Ding, Chao and Bian, Mouxiao and Chen, Jiayuan and Pang, Wenrao and Chen, Ruiyao and Peng, Xinwei and Lu, Renjie and Ren, Sijie and others},
  journal={arXiv preprint arXiv:2511.14439},
  year={2025}
}

@article{zhou2025automating,
  title={Automating expert-level medical reasoning evaluation of large language models},
  author={Zhou, Shuang and Xie, Wenya and Li, Jiaxi and Zhan, Zaifu and Song, Meijia and Yang, Han and Espinoza, Cheyenna and Welton, Lindsay and Mai, Xinnie and Jin, Yanwei and others},
  journal={npj Digital Medicine},
  year={2025},
  publisher={Nature Publishing Group UK London}
}

@article{jarrassier2026can,
  title={Can AI generate safe anaesthesia plans? A comparative evaluation of three large language models on 100 synthetic cases},
  author={Jarrassier, Audrey and de Saint-L{\'e}ger, Frederik Belot and de Rocquigny, Ga{\"e}l and Ari{\`e}s, Philippe and Joseph, Alexandre and Gojon, Yann and Jacques-Sebastien, Clo{\'e} and Riff, Jean-Cl{\'e}ment and Klack, Floriane and Noel, Alexandre and others},
  journal={Anaesthesia Critical Care \& Pain Medicine},
  pages={101769},
  year={2026},
  publisher={Elsevier}
}

\end{document}